# AN ENTITY-DRIVEN RECURSIVE NEURAL NETWORK MODEL FOR CHINESE DISCOURSE COHERENCE MODELING

Fan Xu, Shujing Du, Maoxi Li  and Mingwen Wang

School of Computer Information Engineering, Jiangxi Normal University
Nanchang 330022, China

## ABSTRACT

*Chinese discourse coherence modeling remains a challenge taskin Natural Language Processing field.Existing approaches mostlyfocus on the need for feature engineering, whichadoptthe sophisticated features to capture the logic or syntactic or semantic relationships acrosssentences within a text.In this paper, we present an entity-drivenrecursive deep modelfor the Chinese discourse coherence evaluation based on current English discourse coherenceneural network model. Specifically, to overcome the shortage of identifying the entity(nouns) overlap across sentences in the currentmodel, Our combined modelsuccessfully investigatesthe entities information into the recursive neural network freamework.Evaluation results on both sentence ordering and machine translation coherence rating task show the effectiveness of the proposed model, which significantly outperforms the existing strong baseline.*

## KEYWORDS

*Entity, Recursive Neural Network, Chinese Discourse, Coherence*

## 1. INTRODUCTION

Discourse Coherence Modeling (DCM) aims to evaluate a degree of coherence among sentences within a discourse or text. It is considered one of the key problems in Natural Language Processing (NLP) due to its wide usage in many NLP applications, such as statistical machine translation[1], discourse generation[2][3][4], text automation summarization[3][5][6], student essay scoring[7][8][9].

In general, a coherent discourse generally has many similar components (lexical overlap or coreference) across sentences within a text, while incoherent discourse is the other one. Therefore, the traditional cohesion theory of Centering[10] driven and entity-based model[11][12][13][14]was proposed to capture the syntactic or semantic distribution of discourse entities (nouns) between two adjacent sentences in a text. Thereafter, many extension works were presented such as Feng and Hirst[15]'s multiple ranking model, Lin et al.[16]'s discourse relation-based approach, Louis and Nenkova[17]'s syntactic patterns-based model. However, the potential issue of the existing traditional coherence models need feature engineering, which is a time-consuming job.

In order to overcome the limitation of feature engineering issue, modern research tries to use neural network to extract the syntactic or semantic representation of a sentence automatically. Li et al.[18]proposed neural deep model to deal with English discourse coherence evaluation. However, their discourse coherence model only focuses on the distributed representation for sentences, and did not consider the entity (nouns) distribution across sentences. In fact, the entities can be overlapped between two adjacent sentences, and are good insight to capture the coherence between adjacent two sentences as mentioned in traditional entity-based method. Therefore, we successfully integrate this kind of information into current recursive neural network framework. Evaluation results on both sentence ordering and machine translation coherence rating task show the effectiveness of the proposed model, which significantly outperforms the existing strong baseline.

Therefore, this paper tries to answer the following three questions:

(1) Can the current English discourse coherence models (traditional or neural method) work for Chinese discourse coherence evaluation task?
(2) Can the traditional entity based model be integrated into current deep model?
(3) Which kind of word embedding works better for Chinese discourse coherence evaluation?

The rest of this paper is organized as follows. Section 2 reviews related work on discourse coherence modeling. Section 3 introduces the framework of our entity-driven recursive neural network based Chinese discourse coherence model. Section 4 describes the experiment results and detailed analysis. Finally, some conclusions are drawn in Section 5.

## 2. RELATED WORK

In this section, we describe the related work for discourse coherence modeling from traditional and neural network modes, respectively.

### 2.1. TRADITIONAL COHERENCE MODEL

The task of DCM was first introduced by Foltz et al.[19]. They formulated the discourse coherence as a function of semantic relatedness between two adjacent sentences within a text, and employed a vector-based representation of lexical meaning to compute the semantic relatedness. Since then, many supervised approaches to DCM, such as the entity-based model[11][12][13][15], discourse relation-based model[16], syntactic patterns-based model[17], co reference resolution-based model[20][21], content-based model via Hidden Markov Model (HMM)[3][22] and cohesion-driven based model[23] have been proposed in literature.

To be more specific, Barzilay and Lapata[11][12] presented an entity-based model to capture the distribution of discourse entities between two adjacent sentences within a text. As an extensive work of entity-based approach, Lin et al.[16] explored the function of discourse relations to revise the entity and to catch the behavior of discourse relation transfer among sentences. In addition, Feng and Hirst[15] showed that multiple ranking instead of pair wise ranking was effective for the DCM.

Differently, Louis and Nenkova[17] explored the function of syntactic structure in the DCM. Besides, Iida et al.[20] and Elsner et al.[21] demonstrated the importance of the usage of co reference resolution. In addition, Barzilay et al.[3] and Elsner et al.[22] showed that an Hidden Markov Model (HMM)-based content model can be used to capture the topic's transfer from the first sentence to the end sentence of a text, where topics were formulated as hidden states and sentences were treated as observations. Still, a potential issue of the HMM model is its domain-dependent mechanism. Also, Xu et al.[23] explored the impact of Halliday[24]'s Theme Structure Theory (TST) in English discourse coherence modeling. Their model shows the importance of the theme structure, a cohesion theory of Halliday's systemic-functional grammar, to DCM, and the appropriateness of theme and co reference based filtering mechanism.

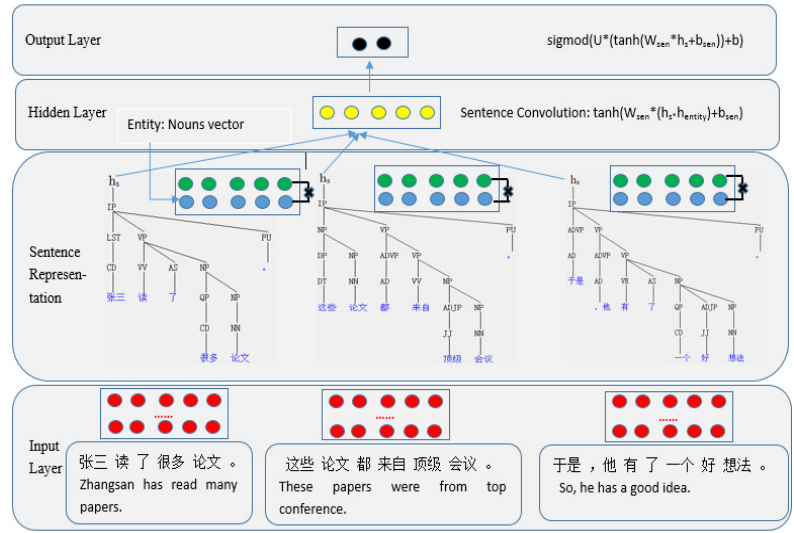

Figure 1: The framework of entity-driven recursive model for Chinese Discourse Coherence Modeling.

## 2.2. NEURAL COHERENCE MODEL

Recently, Li et al.[18] presented a neural deep model for English discourse coherence modeling. They demonstrated the effectiveness of both recurrent and recursive neural network (RNN) model for English situation.

However, as mentioned in the Section 1, their model did not consider the entity (nouns) distribution or entity overlap across sentences within a text. In fact, the entity overlap between two adjacent sentences indicates logical or semantic coherence for para text. Therefore, we successfully integrate these information into their model.

## 3. ENTITY-DRIVEN RNN COHERENCE MODEL

In this section, we describe our entity-driven RNN Chinese discourse coherence model.

### 3.1. FRAMEWORK

Figure 1 shows the entity-driven recursive deep model for Chinese discourse coherence modeling. Our deep model is based on Li et al.[18]'s English discourse coherence framework. On comparison, their model doesn't intensify the effectiveness of entities across each sentences in a text. Therefore, we successfully integrate the entities into current recursive neural network model.

### 3.2. SENTENCE REPRESENTATION

For the word-level representation, each word in a sentence can be represented by using a vector representation (or word embedding), and are able to capture the semantic meanings through toolkit, e.g. word2vec[1] or Glove[2]. More specifically, the word of a sentence can be represented using a specific vector embedding $e_w=\{e_w^1, e_w^2, ..., e_w^K\}$, where $K$ denotes the dimension of the word embedding.

---

[1] http://code.google.com/p/word2vec/
[2] http://nlp.stanford.edu/projects/glove/

For the sentence-level representation, as shown in Figure 1, the vector representation for the whole sentence is computed as a representation for each parent node based on its immediate children recursively in a bottom-up fashion until reaching the root of the tree. Concretely, for a given parent $p$ in the tree and its two children $c_1$(associated with vector representation $h_{c1}$) and $c_2$(associated with vector representation $h_{c2}$), standard recursive network calculates $h_p$for $p$ as follows:

$$h_p = f(W_{Recursive} \cdot [h_{c1}, h_{c2}] + b_{Recursive}) \tag{1}$$

where $[h_{c1}, h_{c2}]$ refers to the concatenating vector for children vector $h_{c1}$ and $h_{c2}$; $W_{Recursive}$ *is a k\*2K matrix and $b_{Recursive}$ is the K\*1 bias vector; f(.) is tanh function.*

## 3.3. ENTITY-DRIVEN SENTENCE CONVOLUTION

The framework treats a window of sentences as a clique $C$(sliding windows of $L$ sentences)and associates each clique with a tag $y_c$ that takes the value 1 if coherent, and 0 otherwise. As shown in Figure 1, each clique $C$ takes as input a *(L\*K)\*1*vector $h_c$ by concatenating the embedding of all its contained sentences. The hidden layer takes as input $h_c$ and performs the convolution using a non-linear tanh function. The concatenating output vector for hidden layers, defined as $q_c$, can therefore be rewritten as:

$$q_c = f(W_{sen} * (h_c * h_{entity}) + b_{sen}) \tag{2}$$

where $W_{sen}$ is a $H*(L*K)$ dimensional matrix and $b_{sen}$ is a $H*1$ dimensional bias vector; $H$ refers to the number of neurons in the hidden layer.

### 3.3.1. ENTITY-DRIVEN MECHANISM

Firstly, we conduct vector summation operation for each nouns' word embedding to generate $h_{entity}$ formulated as:

$$H_{entity} = ew_{NN1} \oplus ew_{NN2} \oplus \ldots \ldots \oplus ew_{NNk} \tag{3}$$

Then, we conduct element wise multiplication operation between $h_c$ and $h_{entity}$.

The value of the output layer can be formulated as:

$$P(y_c = 1) = \text{sigmod}(U^T q_c + b) \tag{4}$$

where $U$ is an $H*1$ vector and $b$ denotes the bias; $y_c$ with value 1 means the text is coherent, and 0 otherwise.

Therefore, the total coherence score for a given document is the probability that all cliques within the document are coherent, which is given by:

$$S_d = \prod_{C \in d} p(y_c = 1) \tag{5}$$

Finally, we can determine whether a text is coherent according to the value of their coherence score.

## 3.4. TRAINING AND OPTIMIZATION

The cost function for the model is given by:

$$J(\Theta) = \frac{1}{M} \sum_{C \in trainset} H_0 + \frac{Q}{2M} \sum_{\theta \in \Theta} \theta^2 \qquad (6)$$

$$H_0 = -y_c \log[p(y_c = 1)] - (1 - y_c) \log[1 - p(y_c = 1)] \qquad (7)$$

where $\Theta = [W_{Recursive}, W_{sen}, U_{sen}]$; M denotes the number of training samples.

We adopt the widely applied optimization diagonal variant of AdaGrad (Duchi et al.[25]) to optimize the loss function.

## 4. EXPERIMENTS

In this section, we demonstrate the effectiveness of our discourse coherence model through both sentence ordering and machine translation coherence rating tasks. The former aims to discern an original text from a permuted ordering of its sentences, while the latter aims to discern a human or reference translation from automatically machine generated translation.

### 4.1. DATASET

**Sentence Ordering Dataset:** We select documents for Chinese Treebank 6.0 from Linguistic Data Consortium (LDC) with catalog number LDC2007T36 and ISBN1-58563-450-6. We select the 100 documents from chtb_2946 to chtb_3045 as our training dataset, and 100 documents from chtb_3046 to chtb_3145 as our testing dataset. The sentences in each source file will be permutated at most 20 times. The total number of testing texts is 1027.The average number of sentence are 10.33 and 13.56 for training set and testing set, respectively. In the evaluation, we consider the original texts are more coherent (positive instances) than the permuted ones (negative instances).

**Machine Translation Dataset:** Similarly**,** we extract documents for NIST Open Machine Translation 2008 Evaluation (MT08) Selected Reference and System Translations from Linguistic Data Consortium (LDC) with catalog number LDC2010T01 and ISBN1-58563-533-2. Therein, the English-to-Chinese language pairs have 127 documents with 1830 segments, output from 11 machine translation systems. The average number of sentence are 13.38 and 13.39 for training and test set, respectively. In evaluation, we consider the human or reference translation texts are more coherent than the machine generated one.

### 4.2. EXPERIMENTAL SETTINGS

**Initialization:** Similar to Li et al.[18], the parameter $W_{sen}$, $W_{recursive}$ and $h_0$ are initialized by randomly drawing from the uniform distribution. The number of hidden layer $H$ is set to 100.Learning rate in the optimization process is set to 0.01, and batch size is set to 20. Differently, word embedding *{e}* for Chinese are trained using word2vec and Glove respectively. The dimension for word embedding is 50 or 100. The window size L is 3 or 5.

**Evaluation Metric:** We report system's performance using accuracy, which is the ratio of the number of the selected original text/translation document divided by the total number of texts/translation document.

**Baseline System 1:** Entity graph based model[14] which has been demonstrated as a simple but effective implementation of the entity-based coherence model. We re-implement their method in this paper based on publicly available code[3].

**Baseline System 2:** Another baseline, Li et al.[18]'s recursive neural model, which did not consider the entity transition information.We transplant there English discourse coherence framework to Chinese situation. Furthermore, we successfully integrate the entity information into their deep model.

In addition, we employ Stanford parser[4]to generate sentence-level constitute parser tree and generate the part-of-speech to get the entities (nouns) occur in each sentence, and use utility ICTCLAS[5] to conduct Chinese word segmentation.

### 4.3. EXPERIMENT RESULTS

In this section, we report the experiment results for the Chinese discourse coherence modeling on both sentence ordering and machine translation coherence rating task.

#### 4.3.1. RESULTS ON SENTENCE ORDERING

Table 1 shows the performance of our entity-driven deep model using different windows size, different dimension, and with different type of word embedding.

Table 1: The performance under different settings on sentence ordering task.

|  | dimension=50 | | dimension=100 | |
| --- | --- | --- | --- | --- |
|  | Window size | | | |
|  | 3 | 5 | 3 | 5 |
| Glove | 56.03 | 49.37 | **65.67** | 52.47 |
| word2vec | 57.52 | 48.68 | 65.56 | 53.50 |

As shown in Table 1, it shows that:

**(1)Dimension**

Generally speaking, the performance increases with the increment of the dimension. In fact, the larger the dimension, the more representative ability it is.

**(2) Window size**

The performance decreases with the increment of the window size, and the best performance yields at the window size with 3. It is mostly caused by the local entity distribution characteristic demonstrated by Barzilay and Lapata[11][12],Guinaudeau and Strube[14]. As the increment of the number of the window size, the entity co-occurance decreases accordingly.

---

Table 2, below, lists the performance comparision among our model and the current baseline model (traditional model and neural network model).

Table 2: Performance comparison among different coherence model on sentence ordering task; Performance that is significantly superior to baseline systems (p<0.05, using paired t-test for significance) is denoted by *.

| | |
|---|---|
| Entity graph based model[14] | 67.78 |
| Li et al.[18]<br>(Glove word embedding) | 65.67 |
| Li et al.[18]<br>(word2vec word embedding) | 65.56 |
| Our combined model | **67.16*(Li's model)** |

It shows that our combined model significantly outperforms the current deep model for Chinese discourse coherence modeling, which demonstrates the effectiveness and importance of the entity distribution across sentences. Interestingly, the traditional entity based model also works for Chinese discourse coherence evaluation, which doesn't work fine for English situation. This is mostly caused by the entity distribution are obvious in Chinese discourse than in English text.

### 4.3.2. RESULTS ON MACHINE TRANSLATION COHERENCE RATING

Table 3, below, lists the performance of our model and the baseline model.

Table 3: Performance comparison among different coherence model on machine translation coherence rating task with dimension equals to 100; Performance that is significantly superior to baseline systems (p<0.05, using paired t-test for significance) is denoted by *.

| | |
|---|---|
| Entity graph based model[14] | 68.50 |
| Li et al.[18]<br>(Glove word embedding) | 70.08 |
| Li et al.[18]<br>(word2vec word embedding) | 68.50 |
| Our combined model | **72.44*** |

In fact, discourse coherence evaluation for the machine translation task is more common than the sentence ordering task evaluation. As the results show in Table 3, again, our model significantly outperforms the current model. Also, our model significantly outperforms the traditional entity-based model. It is mostly caused by the entity distribution is not obvious in the text generated by the machine. But the entity (nouns) information still can be integrated into current recursive neural network model.

## 5. CONCLUSIONS AND FUTURE WORK

In this paper, we present an entity-driven recursive deep model for Chinese discourse coherence modeling. We successfully integrate the entities across each sentence into current recursive neural framework. Evaluation results on both sentence ordering and machine translation coherence rating task show the effectiveness of the proposed model. Our future work is to integrate the co reference mechanism into current combined recursive neural network model, together with other coherence evaluation task.

**ACKNOWLEDGEMENTS**

The authors would like to thank the anonymous reviewers for their comments on this paper. This research was supported bythe National Natural Science Foundation of China under Grant No.61402208, No.61462045, No.61462044, No.61662030, the Natural Science Foundation and Education Department of Jiangxi Province under Grant No. 20151BAB207027 and GJJ150351, and the Research Project of State Language Commission under Grant No.YB125-99.

## AUTHORS

**Fan Xu** holds a Doctoral Degree (Ph.D.) in Computer Science from Soochow University, China. His areas of research interest includes Natural Language Processing, Chinese Information Processing, Discourse Analysis, and Speech Recognition. At present he is working as Lector, School of Computer Information Engineering, Jiangxi Normal University, China. He is member of various professional bodies including ACL, IEEE, and ACIS.

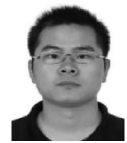

**Shujing Du**is a Master of Computer Science of Jiangxi Normal University, China. Her research interest includes Natural Language Processing, Chinese Information Processing, Discourse Analysis

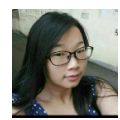

**Maoxi Li** holds a Doctoral Degree (Ph.D.) in Computer Science from Chinese Academy of Sciences. His areas of research interest includes Machine Translation and Natural Language Processing. At present he is working as Associate Professor, School of Computer Information Engineering, Jiangxi Normal University, China.

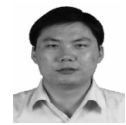

**Mingwen Wang** holds a Doctoral Degree (Ph.D.) in Computer Science from Shanghai Jiaotong University,      China. His areas of research interest includes Machine Learning, Information Retrieval, Natural  Language Processing, Image Processing, and Chinese Information Processing. At present he is working as Professor, School of Computer Information Engineering, Jiangxi Normal University, China. He is member of various professional bodies including ACL, IEEE, CCF, and ACIS.

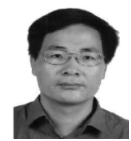